\documentclass{sig-alternate}
\usepackage{subfigure}
\usepackage{amssymb,amsmath}
\usepackage{graphicx}
\begin{document}
%
\title{Coarse-to-Fine Classification via Parametric
and Nonparametric Models for Computer-Aided Diagnosis}
\author{Meizhu Liu ~~~~~~~~~~~~~~~~~~~~~~~~~~~~~~~~ Yahoo Labs, New York, NY 10018\\ 
Le Lu ~~~~~~~~~~~~~~~ National Institutes of Health, Bethesda, MD 20892\\
Xiaojing Ye ~~~~~~~~~~~~~~~~ Georgia State University, Atlanta, GA 30303\\
Shipeng Yu  ~~~~~Siemens Medical Solutions Inc, Malvern, PA 19355}
\maketitle
\vspace{-5mm}
\begin{abstract}
Classification is one of the core problems in Computer-Aided
Diagnosis (CAD), targeting for early cancer detection using 3D
medical imaging interpretation. High detection sensitivity with
desirably low false positive (FP) rate is critical for a CAD system
to be accepted as a valuable or even indispensable tool in
radiologists' workflow. Given various spurious imagery noises which
cause observation uncertainties, this remains a very challenging
task. In this paper, we propose a novel, two-tiered coarse-to-fine
(CTF) classification cascade framework to tackle this problem. We
first obtain classification-critical data samples (e.g., samples on
the decision boundary) extracted from the holistic data
distributions using a robust parametric model (e.g.,
\cite{Raykar08}); then we build a graph-embedding based
nonparametric classifier on sampled data, which can more accurately
preserve or formulate the complex classification boundary. These two
steps can also be considered as effective ``sample pruning'' and
``feature pursuing + $k$NN/template matching'', respectively. Our
approach is validated comprehensively in colorectal polyp detection
and lung nodule detection CAD systems, as the top two deadly
cancers, using hospital scale, multi-site clinical datasets. The
results show that our method achieves overall better
classification/detection performance than existing state-of-the-art
algorithms using single-layer classifiers, such as the support
vector machine variants \cite{Wang08}, boosting \cite{Slabaugh10},
logistic regression \cite{Ravesteijn10}, relevance vector machine
\cite{Raykar08}, $k$-nearest neighbor \cite{Murphy09} or spectral
projections on graph \cite{Cai08}.
\end{abstract}
%


\section{Introduction}
\label{sec:intro}

Colon cancer and lung cancer are the top two leading causes of
cancer deaths in western population. Meanwhile, these two cancers
are also highly preventable or ``curable'' if detected early. Image
interpretation based cancer detection via 3D computer tomography has
emerged as a common clinical practice, and many computer-aided
detection tools for enhancing radiologists' diagnostic performance
and effectiveness are developed in the last decade
\cite{Ginneken10,Murphy09,Ravesteijn10,Slabaugh10,Wang08}. The key
for radiologists to accept the clinical usage of a computer-aided
diagnosis (CAD) system is the high detection sensitivity with
reasonably low false positive (FP) rate per case.

CAD system generally contains two stages: {\em Image Processing} as
extracting sub-volumes of interest (VOI) by heuristic volume
parsing, and informative feature attributes describing the
underlying (cancerous) anatomic structures; {\em Classification} as
deciding the class assignment (cancer, or non-cancer) for selected
VOIs by analyzing features. VOI selection is also called {\em
candidate generation}, or CG, to rapidly identify possibly anomalous
regions with high sensitivity but low specificity, e.g., $>100$
candidates per scan with $1\sim 2$ true positives. Then dozens or
hundreds of heterogeneous image features can be computed per VOI, in
domains of volumetric shape, intensity, gradient, texture and even
context \cite{Ginneken10,Murphy09,Ravesteijn10,Slabaugh10,Wang08}.
Last, the essential goal for classification is to achieve the best
ROC (Receiver Operating Characteristic) or FROC (Free-Response
Receiver Operating Characteristic) analysis on testing dataset, to
balance the criteria of sensitivity and specificity, given VOIs and
associated features.

This paper mainly focuses on the classification aspect of CAD. We
propose and comprehensively evaluate a novel coarse-to-fine
classification framework. The method consists of the following two
steps, in both training and testing. (1) {\em Sample Pruning:}
Parametric classification models (e.g., logistic regression
\cite{Ravesteijn10}, boosting \cite{Slabaugh10}, support/relevance
vector machines \cite{Bi03,Raykar08}) are trained on the complexly
distributed datasets as coarse, distribution-level classification.
The goal is not to assign class labels, but to prune data samples to
select more ``classification-critical'' candidates, which are
expected to preserve the decision boundary in the high dimensional
feature space (thus vast numbers of samples lying far from
classification boundary are discarded \footnote{This is related with
using nearest neighbor analysis to find data samples either near the
decision boundary \cite{Tang06} or in local neighborhoods
\cite{Zhang06}, then training SVM classifiers on reduced or
clustered datasets. However we perform sample pruning by selecting
data upon their classification scores/confidences of a learned
parametric model that is well studied, more robust and stable,
compared with nearest neighbor (NN) clustering method, especially in
high dimensional space. For example, the neighborhood size selection
and defining sensible distance measure problems in NN are
non-trivial.}). (2) {\em Feature Pursuing + $k$NN/Template
Matching:} We first apply feature selection and graph embedding
methods jointly to find intrinsic lower dimensional feature subspace
that preserves group-wise data topology, and then employ
nonparametric classifiers for final classification, using $k$NN or
template matching. We argue that more precisely modeling the
intrinsic geometric of decision boundary, by graph embedding and
nonparametric classifiers in a finer level, can potentially improve
the final classification performance. The overall process is
illustrated as follows \vspace{2mm}
\begin{equation*}
\begin{split}
&\mbox{\textbf{\fbox{Samples}}$\rightarrow$\textbf{\fbox{Sample pruning}}
$\rightarrow$\textbf{\fbox{Feature selection}}}\\
&\mbox{$\rightarrow$\textbf{\fbox{Class regularized graph
embedding}}}\\
&\mbox{$\rightarrow$\textbf{\fbox{$k$NN/Template matching}}}\\
\end{split}
\end{equation*} \vspace{1mm}

We applied our proposed framework on colon polyp and lung nodule
detection, using two large scale clinical datasets collected from
multiple clinical sites across continents. Classification in these
two CAD problems is very important, but also challenging due to the
large within-class variations (for polyps/nodules in different
morphological subcategories, spatial contexts and false positives
resulted by various anatomic structures, such as tagged stool,
ileo-cecal valve, extra-colonic finding and rectal catheter or
balloon for colon polyp detection, and pathology, vessel, vessel
junction, fissure, scar tissue and so on for lung nodule detection).
The low-level imagery data were extracted and presented as the
intermediate-level heterogeneous natured features for the
classification task (as special cases of image based object
recognition). The results show that the proposed framework
significantly outperformed the baseline CAD system using the same
set of input image features, and compared favorably with other
state-of-the-arts.


The rest of the paper is organized as follows. In Section
\ref{sec:RVMMIL} we present (data) sample pruning using a linear
parametric model of Relevance Vector Machine Multiple Instance
Learning (RVMMIL) \cite{Raykar08}. Section \ref{sec:DimRed}
describes the Maximum Relevance Minimum Redundancy (MRMR) based
feature selection and our modified graph embedding method for
stratified optimization of dimension reduction and manifold
projection. This is followed by $k$ nearest neighbor ($k$NN) voting
and $t$-center \cite{Vemuri10} based template matching techniques
for classification in Section \ref{sec:NC}. Integrating sparsity
into graph embedding strategy is also addressed in section
\ref{sec:DimRed}. Then we perform extensive experimental evaluation
using our coarse-to-fine classification diagram on both colon polyp
and lung nodule classification applications in Section
\ref{sec:exp}. Finally we conclude the paper in Section
\ref{sec:con} with discussions .

\section{Sample Pruning using Parametric RVMMIL}
\label{sec:RVMMIL}

We start by developing a ``coarse'' classifier for sample pruning
using a parametric model. Considering the specific characteristics
of CAD classification problems, in this paper we use the RVMMIL
approach \cite{Raykar08}.

Relevance vector machine (RVM) is a supervised Bayesian machine
learning approach that estimates the classifier parameters by
maximizing the likelihood in a probabilistic setting. A powerful
variation/extension has been proposed \cite{Raykar08} to integrate
feature selection and handle multiple instance learning (MIL)
problems which is essential for CAD applications. The output of
RVMMIL is a linear logistic regression model on a reduced set of
features, and gives a class prediction with probability or
confidence for any single instance.

In RVMMIL, the probability for an instance $\mathbf{x}_i$ to be positive is
$p(y=1|\mathbf{x}_i)=\sigma(\mathbf{a}'\mathbf{x}_i)$, where $\sigma$ is the logistic function defined as
$\sigma(t) = 1/(1+e^{-t})$ and $\mathbf{a}'\mathbf{x}_i$ is the linear dot-product between data feature vector $\mathbf{x}_i$ and model coefficient vector $\mathbf{a}$. Therefore, the probability for a bag or set $\mathcal{X}=\{\mathbf{x}_i\}$ to be positive is $p(y=1|\mathbf{x})=1-\prod_{\mathbf{x}_i\in \mathcal{X}}(1-p(y=1|\mathbf{x}_i))$.
Given the training dataset $T=(\mathcal{X},\mathbf{y})$, $\mathcal{X}$ is the set
of training bags of multiple instances with label $\mathbf{y}$. The RVMMIL
utilizes the \textit{maximum a-posterior} (MAP) estimate based on $T$ to find
the optimal parameter $a$ such that
\begin{equation}
\begin{split}
a&=\arg\max_{\tilde{a}} p(\tilde{a}|T)=\arg\max_{\tilde{a}}p(T|\tilde{a})p(\tilde{a})\\
&=\sum_{i}\mathbf{y}_i\log p_i+(1-\mathbf{y}_i)\log (1-p_i)+\log p(\tilde{a}),\label{eqn:logistic}
\end{split}
\end{equation}
where $ p_i=p(\mathbf{y}_i=1|\mathbf{x}_i,\tilde{a})$ and $p(\tilde{a})$ is the prior which can be assumed to be Gaussian. In this
case, (\ref{eqn:logistic}) can be easily solved using Newton-Raphson method
\cite{Raykar08}. For more details, we refer the readers to \cite{Raykar08}.

In our coarse-to-fine classification model, RVMMIL is used as the coarse-level
cascade classifier for sample pruning, i.e., we will remove samples which are not likely to be positive, i.e. $p(y=1|\mathbf{x}_i)<\hat{\rho}$. This step can prune
massive amount of negatives, without hurting much sensitivity by choosing a
balanced $\hat{\rho}$. The retained data samples $p(y=1|\mathbf{x}_i)\geq\hat{\rho}$ are either true positives (at high recall) or ``hard'' false positives lying near
the classification boundary which largely impact the final classification
accuracy. Note that other classifiers with faithful class confidence estimates,
such as boosting \cite{Slabaugh10} and regularized SVM \cite{Wang08}, are also
applicable.

\section{Feature Pursuit via Selection \& Graph Embedding}\label{sec:DimRed}

The basic idea of feature pursuit is to estimate intrinsic, lower
dimensional feature subspace of data for nonparametric
classification, while preserving generative data-graph topology.
This is the key to achieve superior classification performance with
simple nonparametric classifiers. In the proposed framework it
consists of two steps: supervised feature selection and class
regularized graph embedding.

\subsection{Feature Selection}\label{sec:MRMR}

Feature selection, also as known as variable selection, is a machine learning
scheme to search and extract a subset of relevant features so that a desirable
objective of model complexity/effectiveness can be optimized. It essentially has exponential combinatorial complexity in feature cardinality, if doing exhaustive search. By applying feature selection, only a compact subset of highly relevant features is retained, to simplify the later graph embedding or feature projection process and make it more effective. There are many feature selection techniques in the literature \cite{Boutemedjet09,Boutsidis08,Cai10,Guyon03,He05,Liu05,Wolf55,Zhao07}.
In this work, we use Maximum Relevance Minimum Redundancy (MRMR) feature selection \cite{Peng05}, which can give a very good representative feature set with a fixed number of selected features, or the least amount of relevant features to achieve the same accuracy level of representation. More importantly, MRMR is also very efficient in computation and storage.

\begin{figure}[h]
\begin{center}
\subfigure[]
{
 \epsfig{file=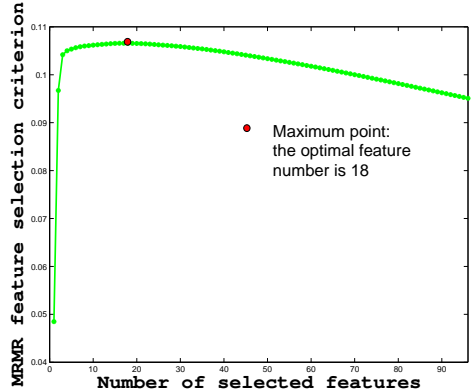, width=0.85\linewidth}
\label{fig:colonFeaSelectionScore}
}
\subfigure[]
{
\epsfig{file=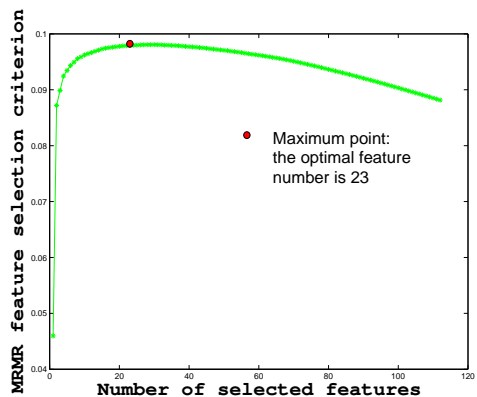, width=0.85\linewidth}
\label{fig:lungFeaSelectionScore}
}
\end{center}
   \caption{The number of selected features versus the MRMR feature selection
criterion in Eq. (\ref{eqn:MRMR}) on colon polyp \subref{fig:colonFeaSelectionScore}
and lung nodule \subref{fig:lungFeaSelectionScore} datasets.
}
\label{fig:MRMRFeatureSelection}
\end{figure}

The relevance in MRMR is measured using a variant of {\em Pearson coefficient} \cite{Rodgers88}.
For any two variables $f$ and $\tilde{f}$, the {\em Pearson}
coefficient $\gamma$ between them is
\begin{equation}\label{eqn:correlation}
\begin{split}
\gamma(f,\tilde{f})&=\frac{|\mathbf{Cov}(f,\tilde{f})|}{\sqrt{\mathbf{Var}(f)\mathbf{Var}(\tilde{f})}}, \\
\mbox{ where }&\mathbf{Cov}(f,\tilde{f})=\mathbf{E}[(f-\mathbf{E}[f])(\tilde{f}-\mathbf{E}[\tilde{f}])], 
\end{split}
\end{equation}
$\mathbf{E}[\cdotp]$ is the expectation and $\mathbf{Var}(\cdotp)$ represents the variance.
Given a set of features $\mathbb{F}=\{f_i\}$, its MRMR feature subset
$\mathbb{H}$ maximizes the following objective $\kappa$:
\begin{equation}\label{eqn:MRMR}
 \kappa(\mathbb{H},\mathbf{y})= \gamma(\mathbb{H},\mathbf{y})-\gamma({\mathbb{H}}),
\end{equation}
where
\begin{align}
 \gamma(\mathbb{H})&=\frac{1}{m^2}\sum_{f_i,f_j\in \mathbb{H}}\gamma(f_i,f_j),\\
\gamma(\mathbb{H},\mathbf{y})&=\frac{1}{m}\sum_{f_i\in \mathbb{H}} \gamma(f_i,\mathbf{y}),
\end{align}
and $m$ is the total number of elements in $\mathbb{H}$.
Suppose we have selected $\mathbb{H}_{i-1}$, the $i$th feature $f_i$ can be selected by
\begin{equation}
 f_i= \arg\max_{f\in
\mathbb{F}-\mathbb{H}_{i-1}}\gamma(f,\mathbf{y})-\frac{1}{i-1}\sum_{f_j\in
\mathbb{H}_{i-1}}\gamma(f,f_j)
\end{equation}
Then $f_i$ will be added to $\mathbb{H}_{i-1}$ to
form $\mathbb{H}_i$ incrementally. If
$\kappa(\mathbb{H}_{i-1},\mathbf{y})\geq \kappa(\mathbb{H}_{i},\mathbf{y})$, then
$\mathbb{H}_{i-1}$ reaches optimum and the iteration will stop. Using this method,
we select 18 out of 96 features for the colon dataset,
 and 23 out of 120 features for the lung nodule dataset.
The objective plots are shown in Fig. \ref{fig:MRMRFeatureSelection}.

\subsection{Class Regularized Graph Embedding}
\label{sec:graphEmbedding}

Nonparametric classifiers, as nearest neighbor (NN) or ($t$-center \cite{Vemuri10}) template matching (TM), can be flexible and powerful representations for joint classification, clustering and retrieval, but they are also sensitive to high dimensional feature space. In this section, we exploit {\em Class Regularized Graph Embedding} (CRGE) to project data (after feature selection) into an even lower dimensional subspace, where data samples from the same class getting closer and samples from different classes moving apart, to make NN or TM more robust and semantically interpretable, as shown later.

Graph embedding is a special class of dimension reduction method using linear or nonlinear projections. 
Feature projections can be learned in different ways: minimizing the reconstruction error as in principal component analysis (PCA) \cite{Duda00,Jolliffe86}; preserving distances in the original space, e.g. multidimensional scaling (MDS) \cite{Cox08} and ISOMAP \cite{Tenenbaum00}; maximizing class-data separation as linear discriminant
analysis (LDA) \cite{Duda00}, or retaining the linear relationship
between locality neighbors, e.g., neighborhood component analysis (NCA)
\cite{Goldberger04}, locally linear embedding (LLE) \cite{Roweis00}.
We follow {\em the principle that keeps the locality of nearby data and maps
apart data further}, in the graph-induced subspace, which is similar to
Laplacian Eigenmap \cite{Belkin03,Cai07} and Locality Preserving Projection
\cite{He03}.

Given a set of $N$ points
$\mathcal{X} =\{\mathbf{x}_1,\mathbf{x}_2,\cdots, \mathbf{x}_N\}\subset \mathbb{R}^n$, and a symmetric
$N\times N$ matrix $W$ which measures the similarity between all pairs of points
in $\mathcal{X}$. The set $\mathcal{X}$ and matrix $W$ compose a graph
$\mathcal{G}$, with $\mathcal{X}$ as vertices and $W$ as weights of the edges.
The conventional graph embedding method will map $\mathcal{X}$ to
a much lower dimensional space $\mathcal{Y}=\{\mathbf{y}_1,\mathbf{y}_2,\cdots, \mathbf{y}_N\} \subset
\mathbb{R}^{\tilde{n}}$, $\tilde{n}\ll n$. The optimal $\mathcal{Y}$ should
minimize the loss function $L(\mathcal{Y})$ which is defined as
\begin{equation}\label{eqn:energyGE}
L(\mathcal{Y})= \sum_{i,j} \|\mathbf{y}_i-\mathbf{y}_j\|^2W_{ij},
\end{equation}
under some appropriate constraints. This objective function ensures $\mathbf{y}_i$ and $\mathbf{y}_j$ to be close if $\mathbf{x}_i$ and $\mathbf{x}_j$ are close. Though performed well in many applications \cite{Cai07,He03}, the limitation of Eq.
(\ref{eqn:energyGE}) is that it does not penalize the similarity between points
belonging to different classes. One more comprehensive strategy is to simultaneously
maximize the similarity between data pairs of the same class and minimize the
similarity between two points rooted from different classes. In other words, we
optimize on mapping the same class data to proximity subspaces, while projecting
different class data samples to be far apart, explicitly.

The goal of class regularized graph embedding is to find a
mapping $\phi:\mathcal{X}\mapsto\mathcal{Y}$,
such that $\phi$ minimizes the function $E(\mathcal{Y})$ defined as
\begin{equation}\label{eqn:energy}
\begin{split}
&E(\mathcal{Y})= \sum_{i,j\in \mathcal{S}}
\|\mathbf{y}_i-\mathbf{y}_j\|^2W_{ij}-\sum_{i,j\in \mathcal{D}}\|\mathbf{y}_i-\mathbf{y}_j\|^2W_{ij}\,,\\
&\mbox{subject to: }\|\mathcal{Y}\|_F=1.
\end{split}
\end{equation}
where $i,j\in \mathcal{S}$ means $\mathbf{x}_i$ and $\mathbf{x}_j$ belong to the same class, and $i,j\in \mathcal{D}$ means $\mathbf{x}_i$ and $\mathbf{x}_j$ are in different classes. To avoid notation clutter, we
rewrite (\ref{eqn:energy}) and get
\begin{equation}\label{eqn:energy2}
\min \sum_{i,j} \|\mathbf{y}_i-\mathbf{y}_j\|^2W_{ij}H_{ij}\,,
\end{equation}
where $H_{ij}$ is the Heaviside function and
$$
H_{ij} = \left\{ \begin{array}{rr}
 1, &\mbox{ if } i,j \in \mathcal{S}\\
  -1, &\mbox{ if } i,j \in \mathcal{D}
       \end{array} \right..
$$
Various choices of the mapping function $\phi$ have been proposed recently,
e.g. linear mapping, kernel mapping and tensor mapping \cite{Yan06}. We use
linear mapping because of its simplicity and generality \cite{Cai10}. A linear mapping
function $\phi$ is described as
\begin{equation}\label{eqn:philinear}
\mathbf{y} = \phi(\mathbf{x}) = M'\mathbf{x}, \, M\in \mathbb{R}^{n\times\tilde{n}}\,, \tilde{n}\ll n \,.
\end{equation}
Plugging (\ref{eqn:philinear}) into (\ref{eqn:energy2}),
we get
\begin{equation}\label{eqn:energy3}
\begin{split}
&\min_M \sum_{i,j} \|M'\mathbf{x}_i-M'\mathbf{x}_j\|^2W_{ij}H_{ij}\,,\\
&\mbox{subject to: } \|M\|_F=1\,,
\end{split}
\end{equation}
 where $\|\cdotp\|_F$ is the Frobenius norm, and the constraint $\|M\|_F=1$
eliminates the scaling effect. Eq. (\ref{eqn:energy3}) can be solved very quickly
using gradient descent technique along with iterative projections
\cite{Rockafellar70}. The reduced dimension $\tilde{n}$ is determined
when the loss function (\ref{eqn:energy}) is minimized by varying $\tilde{n}$.
Though some other ways are possible.  

The computation of $W$ can be done in the following manners, which correspond to
different dimension reduction methods as LLE \cite{Roweis00},
ISOMAP \cite{Tenenbaum00}, and Laplacian Eigenmap \cite{Belkin03,Cai07}.
\begin{align}
 W(i,j)&= \left\{ \begin{array}{rr}
 1, &\mbox{ if } i,j \in \mathcal{S}\\
 0, &\mbox{ if } i,j \in \mathcal{D}
       \end{array} \right.\,; \label{eqn:W1}\\
W(i,j)&= \exp\{-\alpha \|\mathbf{x}_i-\mathbf{x}_j\|^2\}\,,\alpha>0\,; \label{eqn:W2}\\
W(i,j)&= \exp\{-\alpha (\mathbf{x}_i-\mathbf{x}_j)'A(\mathbf{x}_i-\mathbf{x}_j)\}\,,\nonumber\\
&\alpha>0, A \mbox{ is a PSD
matrix}\,; \label{eqn:W3}\\
W(i,j)&= \mathbf{x}_i'\mathbf{x}_j/\|\mathbf{x}_i\|\|\mathbf{x}_j\|\,. \label{eqn:W4}
\end{align}
Eq. (\ref{eqn:W1}) is the simplest weighting scheme, where $W(i,j)=1$ if and only if $\mathbf{x}_i$ and $\mathbf{x}_j$ belong to the same class. However this scheme might lose information about the affinity between the nodes belonging to different classes. Eq. (\ref{eqn:W2}) is the heat kernel weighting method, which has an intrinsic
connection to the Laplace Beltrami operator on differential functions on a manifold \cite{Belkin01}. Eq. (\ref{eqn:W3}) is related to the Mahalanobis distance
 between two vectors. Eq. (\ref{eqn:W4}) is the dot product weighting scheme, which measures the cosine similarity of the two vectors and is easy to compute.
For our CAD purpose of cancer lesion classification, Eq. (\ref{eqn:W1}) neglects the similarity between negative and positive samples, which invalidates the penalization about the similarity between samples from different classes; Eq. (\ref{eqn:W2}) and (\ref{eqn:W3}) are not suitable because they both use Euclidean or Mahalanobis similar distance assumption, which holds when the data samples lie in a (locally) Euclidean space. From our empirical observation, this assumption does not apply to colon polyp or lung nodule dataset. Furthermore, Eq. (\ref{eqn:W2}) and (\ref{eqn:W3}) bother to tune the parameters $\alpha$ or $A$ which may be sensitive for the similarity calculation. Thus we use (\ref{eqn:W4}) for its appropriateness and computation efficiency.

\begin{figure}[ht]
\begin{center}
\subfigure[]{
\epsfig{file=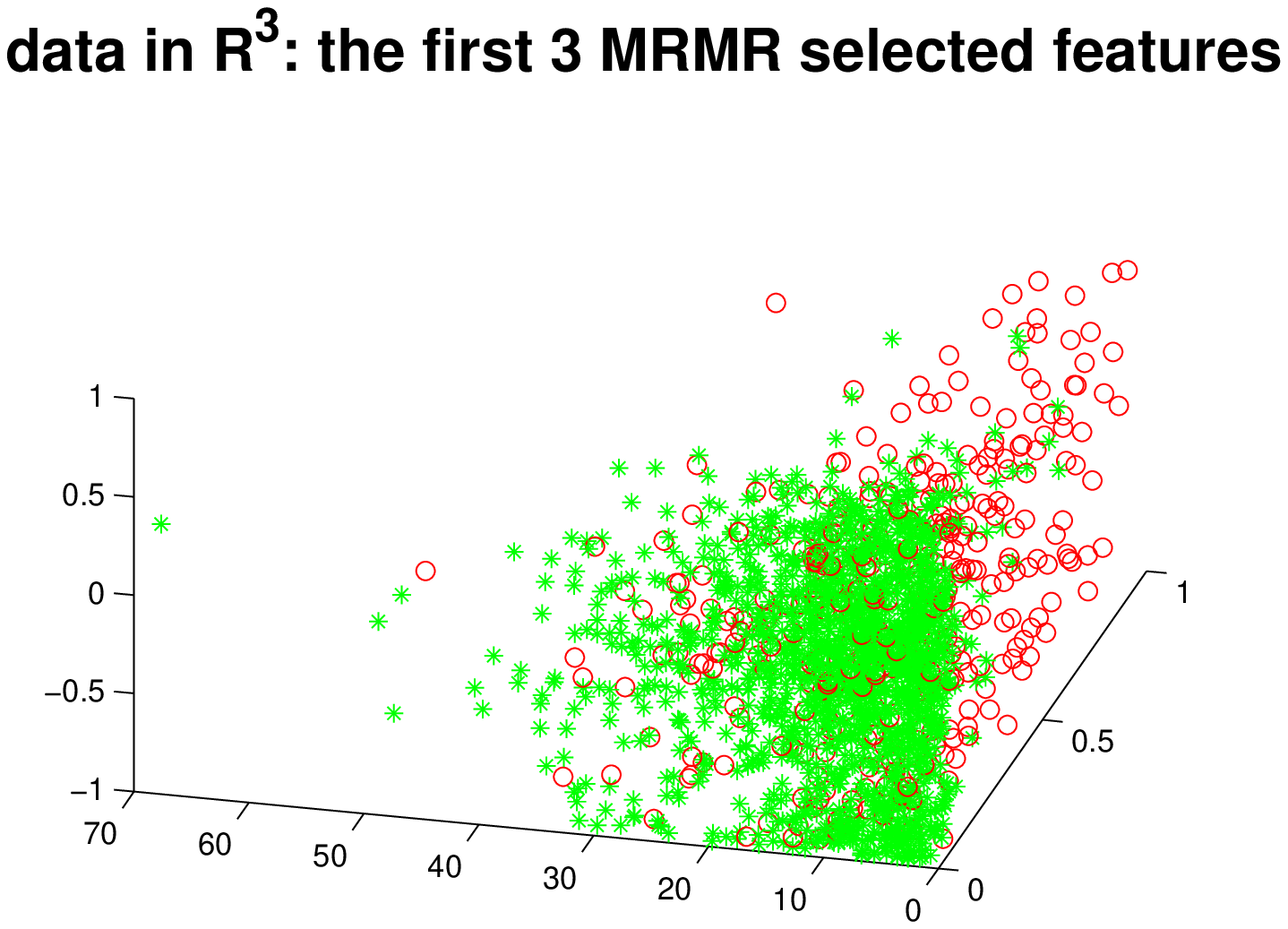, width=0.46\linewidth}
\label{fig:colonBeforeDMFea3}
}
\subfigure[]{
\epsfig{file=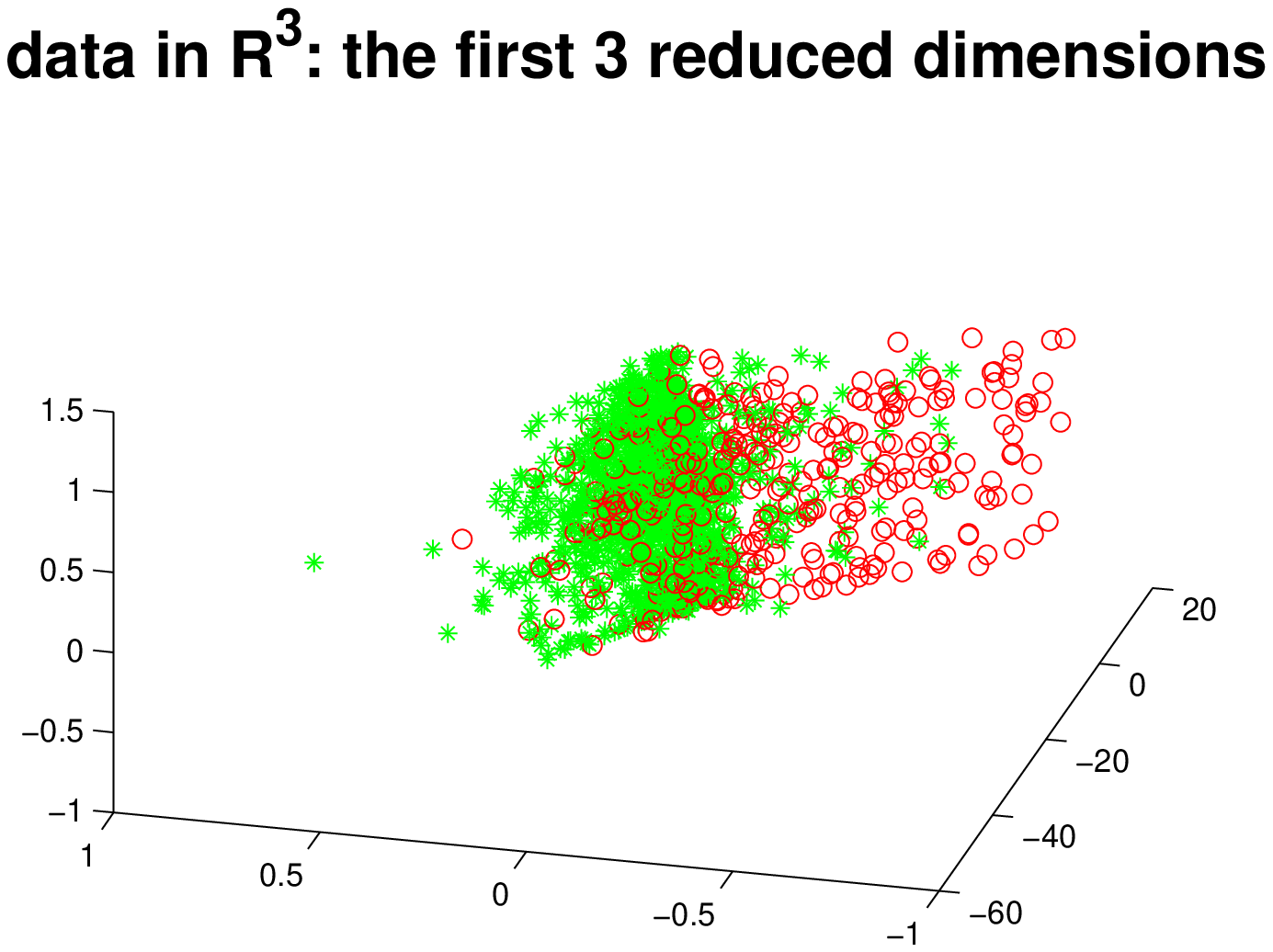, width=0.46\linewidth}
\label{fig:colonAfterDMFea3}
}
\subfigure[]{
\epsfig{file=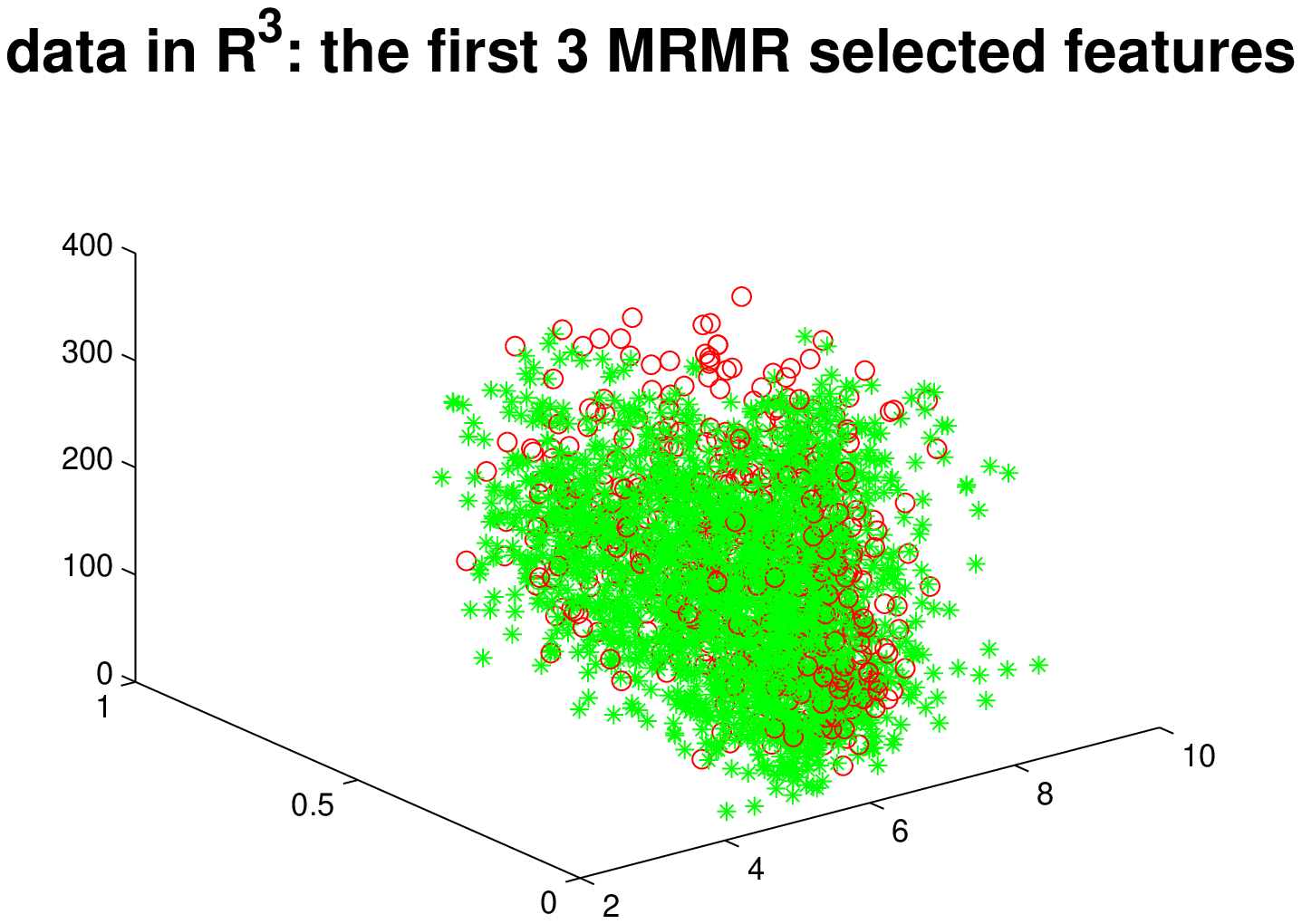, width=0.46\linewidth}
\label{fig:lungBeforeDMFea3}
}
\subfigure[]{
\epsfig{file=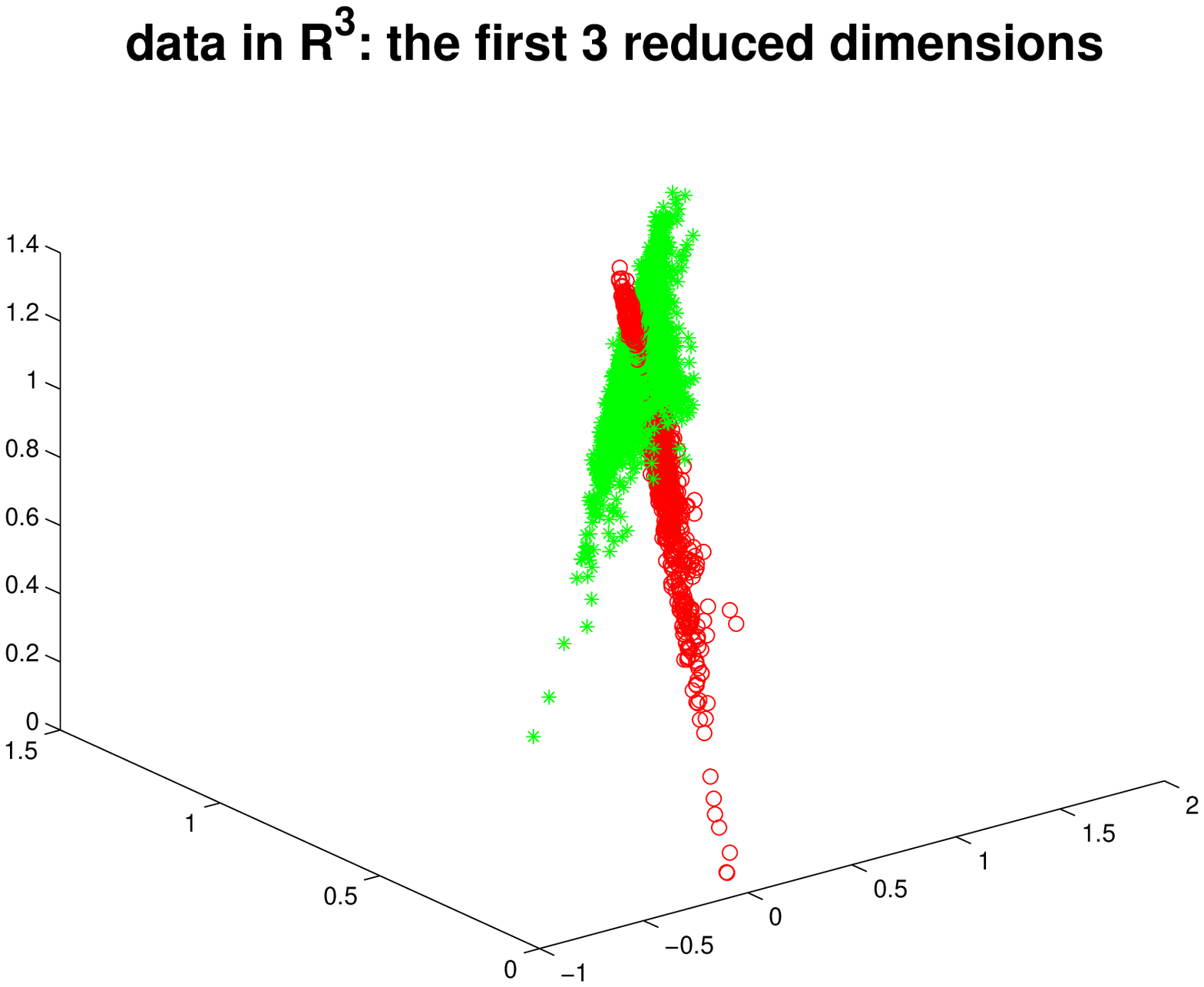, width=0.46\linewidth}
\label{fig:lungAfterDMFea3}
}
\end{center}
\caption{Plot of the data samples (testing) according to the first three
features selected by MRMR \subref{fig:colonBeforeDMFea3}
and the first three dimensions from graph embedding
\subref{fig:colonAfterDMFea3}
on the colon polyp dataset.
Similarly, \subref{fig:lungBeforeDMFea3} and
\subref{fig:lungAfterDMFea3} are illustrated based on the lung nodule dataset.
The dimension coordinates on the figures are not directly
comparable.}
\label{fig:beforeAfterDMFeaCompare}
\end{figure}

The effectiveness of dimension reduction can be evaluated according to several
criteria, e.g., information gain \cite{Cover06}, Pearson coefficients \cite{Rodgers88} and Fisher score \cite{Fisher36}. We validate the effectiveness of our proposed dimension reduction technique using {\em Fisher Score} (FS) \cite{Fisher36} on both polyp colon and lung nodule datasets. The class separability between negatives and positives is measured via Fisher's linear discriminant \cite{Fisher36}. Let the covariance matrices of the negatives and positives be $\Sigma_{-}$ and $\Sigma_{+}$, and the means of the negatives and positives be $\mu_{-}$ and $\mu_{+}$, then the Fisher linear discriminant of the binary classes is \begin{equation}\label{eqn:Fisher}
 s= (\mu_{+}-\mu_{-})'(\Sigma_{+}+\Sigma_{-})^{-1}(\mu_{+}-\mu_{-})\,,
\end{equation}
where the larger $s$ is, the more statistically distinguishable
negative-positive class distributions will be.
CRGE is capable to increase the discriminant between positive and negative
lesions in the projected feature subspaces, visually and numerically.
This is validated on the colon polyp and lung nodule datasets.
For comparison, we plot the first three MRMR selected original features and the
first three projected dimensions after CRGE, on (testing) colon polyp
and lung nodule datasets in Fig. \ref{fig:beforeAfterDMFeaCompare}.
The Fisher (linear discriminant) score for the first three MRMR selected
features on the colon polyp dataset is $0.2725$, and after CRGE, the score
improves to $0.7990$. For the lung nodule dataset, the score increases from
$0.1083$ to $0.6987$, reflecting the impact of CRGE. The numerical results
demonstrate that our class regularized graph embedding technique indeed enlarges the class separability between negative and positive populations, for both datasets. Note that many dimension reduction methods are tested using image data where each dimension is a pixel or voxel, for classification \cite{Cai07,He03} and registration \cite{Hamm10}. As mentioned above, CAD image features are extremely heterogeneous attributes as measuring different nature imaging properties for 3D VOI structures, in different metrics or dimensions.

\subsection{Sparse Graph Embedding}
\label{subsec:sparseGraphEmbedding}

As a companion to the above stratified ``feature pursuing'' strategy of {\em feature selection + graph embedding}, an integrated approach is Sparse (feature) Projections over Graph (SPG) \cite{Cai08, Cai10}. SPG utilizes techniques from graph theory \cite{Chung97} to construct an affinity graph over the data and assumes that the affinity graph is usually sparse (e.g. nearest neighbor graph). Thus the embedding results can be efficiently computed. After this, lasso regression \cite{Efron04} is applied to obtain the sparse basis functions. The data in the reduced subspace is represented as a linear combination of a {\em sparse subset} consisting of the most relevant features, rather than using all features as in PCA, LDA or regular graph embedding. Feature selection and graph embedding based dimension reduction are jointly presented and formulated within the same optimization framework.

The SPG algorithm is described as follows. Given a set of $N$ points
$\mathcal{X} =\{\mathbf{x}_1,\mathbf{x}_2,\cdots, \mathbf{x}_N\}\subset \mathbb{R}^n$, the goal of SPG is to find a transformation matrix $A=(\mathbf{a}_1,\cdots,\mathbf{a}_{\tilde{n}})$
that maps the N points to a set of lower dimensional
points $\mathcal{Y}=\{\mathbf{y}_1,\mathbf{y}_2,\cdots, \mathbf{y}_N\} \subset
\mathbb{R}^{\tilde{n}}$, $\tilde{n}\ll n$. For each $i$, $\mathbf{y}_i(=A'\mathbf{x}_i)$ is the projection of
$\mathbf{x}_i$ onto the lower dimensional space $\mathbb{R}^{\tilde{n}}$. Furthermore, there is a sparsity constraint on each projection $\mathbf{a}_i$, and $\|\mathbf{a}_i\|_0<k$ ($k<n$), where $\|\mathbf{a}\|_0$ is defined as the number of nonzero entries of $\mathbf{a}$. To obtain the optimal projection,
one first needs to create a graph G with affinity matrix W over $\mathcal{X}$, and then minimize the following energy function
\begin{equation} \label{eqn:SPG}
 \begin{split}
 &\min_{\mathbf{a}}\mbox{  }\sum_{i,j}(\mathbf{a}'\mathbf{x}_i-\mathbf{a}'\mathbf{x}_j)^2W_{ij}\\
&\mbox{subject to: }\mathbf{a}'XDX'\mathbf{a}=1\,,\\
&\|\mathbf{a}\|_0\leq k\,,
 \end{split}
\end{equation}
where $X=(\mathbf{x}_1,\mathbf{x}_2,\cdots, \mathbf{x}_N)$, $D$ is a diagonal
matrix and each entry of the diagonal is the sum of the corresponding row of W, i.e., $D_{ii}=\sum_jW_{ij}$. Since Eq. (\ref{eqn:SPG}) is NP-hard, it is split into two steps. The first step introduces the graph Laplacian matrix \cite{Chung97} $L=D-W$, and the optimization function in Eq. (\ref{eqn:SPG}) can be reformulated as
\begin{equation}\label{eqn:SPG2}
 \sum_{i,j}(\mathbf{a}'\mathbf{x}_i-\mathbf{a}'\mathbf{x}_j)^2W_{ij}=\mathbf{a}'XLX'\mathbf{a}
\end{equation}
The solution to (\ref{eqn:SPG2}) with the first constraint in (\ref{eqn:SPG}) leads to
\begin{equation}\label{eqn:SPG3}
 XLX'\mathbf{a}=\lambda XDX'\mathbf{a}.
\end{equation}

Once obtaining the embedding $y_i=\mathbf{a}'\mathbf{x}_i$, lasso regression can be applied to get the sparse transformation according to the following minimization
\begin{equation}\label{eqn:lasso}
\min_{\tilde{\mathbf{a}}}\left(\sum_{i=1}^m(y_i-\tilde{\mathbf{a}}'\mathbf{x}_i)^2+ \beta\|\tilde{\mathbf{a}}\|_1\right).
\end{equation}

After learning the sparse transformation $\mathbf{a}$, we can project all the samples into the lower and more intrinsic dimensional space, in which we can perform classification. SPG, in some sense, integrates the feature selection
and dimension reduction processes, which has been shown to be effective in many applications, such as text clustering \cite{Cai08} and classification on many benchmark machine learning datasets \cite{Cai10}. {\em However, we argue that our stratified approach which prunes non-informative or redundant features from an information-theoretic aspect before graph embedding or feature projection, can simplify the optimization process of graph embedding on a reduced feature set. This strategy may achieve better overall results, compared from the holistic sparsity-constrained graph embedding (as SPG).} The sparse approximation after embedding (i.e., Eq. (\ref{eqn:lasso})) is also suboptimal. In practice, superior classification performances over two hospital scale, clinical datasets are demonstrated using our stratified feature pursuit framework, in later experimental section.

\section{Nonparametric Classification} \label{sec:NC}

After finding the mapping $\phi$ and $\mathcal{Y}$, we will perform unsupervised
clustering on $\mathcal{Y}$ for training negatives and positives separately.
Data samples of the same class are divided into local clusters, where instances from the same cluster are more similar than instances from different clusters. Each cluster is then represented using a template. Based on the $k$NN voting of the cluster templates, each instance in testing is labeled. The explanation of clustering and calculating the templates is detailed in the following section.

\subsection{Clustering \& Templates}

The clustering process is performed according to a recently introduced clustering algorithm, namely total Bregman divergence clustering algorithm \cite{Liu10}. This algorithm utilizes the newly proposed divergence measure first presented in \cite{Vemuri10}. This divergence measure is called total Bregman divergence (tBD) which is based on the orthogonal distance between the convex generating
function of the divergence and its tangent approximation at the second
argument of the divergence. tBD is naturally robust and leads to
efficient algorithms for soft and hard clustering. For more details,
we refer the reader to \cite{Liu10,Vemuri10}.

We employ the total Bregman divergence hard-clustering algorithm
\cite{Liu10} to perform clustering on negative or positive data instances, in
$\mathcal{Y}$ space. Denote that $c_1$ clusters, with the
cluster centers $\{z_{i-}\}_{i=1}^{c_1}$, are obtained for negatives;
and $c_2$ clusters with centers $\{z_{j+}\}_{j=1}^{c_2}$ for positives.
The number of clusters $c$ is chosen to minimize the {\em intra-inter-validity
index} \cite{Ray99}, given by
\begin{equation}
\begin{split}
 \mbox{index}&=\frac{\mbox{intra}}{\mbox{inter}},\\
 \mbox{intra}&=\frac{1}{N}\sum_{i=1}^{c}\sum_{y\in C_i}\|y-z_i\|^2,\\
 \mbox{inter}&=\min_{i,j}\|z_i-z_j\|^2,
\end{split}
\end{equation}
where $C_i$ is the $i$th cluster with center $z_i$. Each cluster is represented
as the {\em tBD} center, termed $t$-center \cite{Liu10,Vemuri10}, which is the $\ell_1$ norm median of
all samples in the corresponding cluster. For example,
if $\{\mathbf{y}_{i}\}_{i=1}^N$ is the set of samples, then its $t$-center $z$ will be
\begin{equation}\label{eqn:tcenterPosNeg}
z=\arg\min_{\tilde{z}}\sum_{i=1}^N\delta_f(\tilde{z},\mathbf{y}_{i}),
\end{equation}
where $\delta_{f}$ is the total Bregamn divergence generated by the convex and differentiable
 generator function $f$, and
\begin{equation}\label{eqn:delta}
\delta_{f}(\mathbf{y}_1,\mathbf{y}_2)=\frac{f(\mathbf{y}_1)-f(\mathbf{y}_2)-\langle \mathbf{y}_1-\mathbf{y}_2,\nabla
f(\mathbf{y}_2)\rangle}{\sqrt{1+\|\nabla
f(\mathbf{y}_2)\|^2}}.
\end{equation}
 Here, if we use
$f(y)=\|y\|^2$, $\delta_{f}$ becomes the total square loss \cite{Liu10,Vemuri10} and the $t$-center in
Eq. (\ref{eqn:tcenterPosNeg}) becomes
\begin{equation}
z=\sum_{i=1}^Na_i\mathbf{y}_{i}\,,\mbox{ where } a_i=\frac{1/\sqrt{1+4\|\mathbf{y}_i\|^2}}{(\sum_j
1/\sqrt{1+4\|\mathbf{y}_j\|^2})}\,.\label{eqn:tcenterPosNeg2}
\end{equation}
After learning the centers as templates, we can predict whether a given sample
is positive or negative, according to the $k$NN voting on the set of trained
positive/negative $t$-centers.

\subsection{Template Matching via $k$NN Voting}

Nearest neighbor voting is a popular nonparametric classifier which has been
studied extensively \cite{Salem10}. Given a test sample $\mathbf{y}_i$, we will find its $k$ nearest neighbors from the $t$-centers. Suppose the neighbors are
$\{z_1,z_2,\cdots,z_k\}$ and the corresponding distance from $\mathbf{y}_i$ to the
neighbors are $\{d_1,d_2,\cdots,d_k\}$.
The distance $d_i$ can be Euclidean distance or the vector angle difference
(Euclidean distance is used in our experiments). We define the empirical
probability of $\mathbf{y}_i$ being positive as $p$($\in [0,1]$), where
\begin{equation}\label{eqn:predict1}
p=\frac{\sum_{(z_j \mbox{ is positive})}1/d_j}{\sum_{(z_l \mbox{ is
negative})}1/d_l + \sum_{(z_j \mbox{ is positive})}1/d_j}\,.
\end{equation}
Based on the $p$ value, we can draw the FROC curve of sensitivity and FP rate per
case for training and testing datasets. Eq. (\ref{eqn:predict1})
is a soft $k$NN voting scheme using the inverse of distance $1/d_i$.
There are other options to calculate $p$, e.g., using the counts of
positive/negative $t$-centers. We argue that using $t$-centers, instead of proximity data samples for $k$NN voting is more robust, given more sparsity and diversity of CAD lesion data distributions.

The number of nearest neighbors $k$ is chosen during the training/validation
stage. Since the optimal $k$ should give our algorithm the possibly highest
performance, we set $k$ to be the one maximizing the Area Under (the FROC) Curve
(AUC) on the training dataset. Additionally, if only a partial range of FROC has
more meaningful impacts on clinical practice (e.g., $FP \in [2,\,4]$ per case),
we can search $k$ to optimize the partial AUC
\begin{equation}
k=\arg \max_{\tilde{k}} \mbox{ AUC}(\mbox{FPrate} \in [2,\, 4]).
\end{equation}

\section{Experimental results}
\label{sec:exp}

Unlike many existing lesion classification systems \cite{Furlan11,Lo03,Pereira06}
which use small datasets, our method is evaluated on representative large scale datasets with great diversity, which are collected from dozens of hospitals across US, Europe and Asia. We perform two important clinical tasks of classifying colonic polyps and lung nodules from 3D CT imagery features. Lung cancer and colon cancer are the two leading deadly cancers in western population.

\subsection{Colon Polyp Detection \& Retrieval}\label{subsec:colon}

{\bf Data:}
The colon polyp dataset contains 134116 polyp candidates obtained from an annotated CT colonography (CTC) database of 429 patients. Each sample is represented by a 96-dimensional computer extracted feature vector, describing its shape, intensity pattern, segmented class-conditional likelihood statistics and other higher level features \cite{Ravesteijn10,Lu11,Slabaugh10,Wang08}. The patients were examined from 12 hospitals via different scanners from Siemens, GE and Philips and under various fecal-tagging imaging protocols. Each patient is scanned in two positions resulting  two (prone and supine) scans. Out of the 134116 samples, there are 1116 positive samples. The goal of classification for the colon dataset is to determine whether a sample is negative (false positive) or positive (true polyp). The CAD sensitivity is calculated at per-polyp level for all actionable polyps $\geq 6mm$ (i.e., polyp is classified correctly at least from one view), and the FP rate counts the sum of two (prone-supine) scans per patient. The colon polyp dataset is split into two parts, training dataset and testing dataset. The training and testing datasets is split at patient level. No data from the same patient is used for both training and testing. Here, we do not employ N-fold cross validation because we intend to keep a portion of data (as our testing dataset) which is always unseen for training. This is practically critical to evaluate the more ``true'' or trustful performance of a clinical product. As a result, the training dataset contains all the instances detected from 216 patients, and the testing dataset includes the other 213 patients.

\begin{figure}[h]
\centering
\subfigure[]{
\epsfig{file=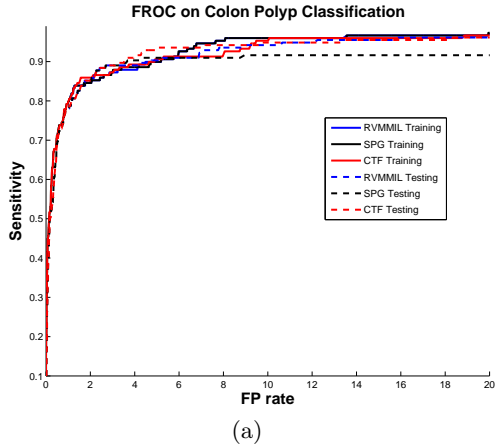, width=0.78\linewidth}
\label{fig:polypMatching}
}
\subfigure[]{
\epsfig{file=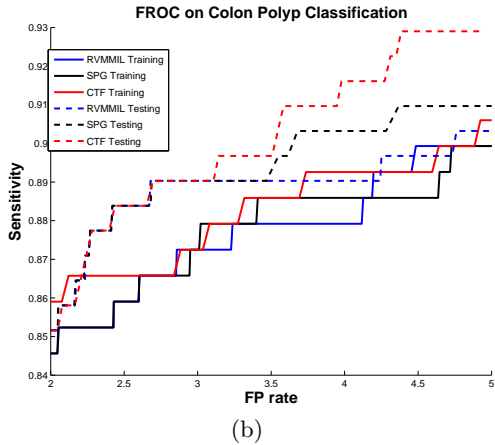, width=0.78\linewidth}
\label{fig:polypMatchingZoomIn}
}
\caption{\subref{fig:polypMatching} FROC comparison of using our proposed CTF method,
single-layer RVMMIL \cite{Raykar08} classifier
and spectral projection on graph (SPG) \cite{Cai08} on classifying the training and testing datasets of colon polyps.
\subref{fig:polypMatchingZoomIn} Zoom in of \subref{fig:polypMatching} for the part of FP rate $\in[2,\,5]$.}
\label{fig:polypMatchingCompare}
\end{figure}

After estimating the parametric RVMMIL model \cite{Raykar08}, we get the probability (classification score) for each candidate to be positive. Then we perform thresholding according to the classification scores. Let the condition on classification scores $p(y=1|\mathbf{x}_i)\geq\hat{\rho}=0.0157$ as a cascade with high-recall, we obtain total $3466$ data samples, pruned from $134116$ polyp candidates on the training dataset. All $554$ true positive lesion instances are contained, along with other ``harder'' negatives having higher classification scores. For fine-level classification, we learn the mapping function $\phi:\mathcal{X}\mapsto\mathcal{Y}$ after feature selection using the pruned dataset, and the $t$-centers are fitted in the reduced $\mathcal{Y}$ feature space for the soft $k$NN classifier. We plot the FROC curves comparing using RVMMIL as a single classifier, using SPG \footnote{We use the code implemented by Dr. Cai Deng http://www.zjucadcg.cn/dengcai/SR/index.html} as a integrated dimension reduction approach, and our two-tiered coarse-to-fine classifier, on training and testing datasets, as shown in Fig. \ref{fig:polypMatchingCompare}.
Fig. \ref{fig:polypMatching} shows the whole FROC curve. Since
the more clinically meaningful region on FROC is when the FP rate is reasonably small, we highlight in the partial-FROC with FP rate $\in[2,\,5]$ and illustrate it in Fig. \ref{fig:polypMatchingZoomIn}.
For validation, the testing results demonstrate that our CTF method can increase the sensitivity of RVMMIL by $2.58\%$ (from 0.8903 to 0.9161) at the FP rate $=4$, or reduce the FP rate by 1.754 (from 5.338 to 3.584) when sensitivity is 0.9097, which are statistically significant for colorectal
cancer detection. It also clearly outperforms other state-of-the-arts, e.g. SPG \cite{Cai08} as shown in Fig. \ref{fig:polypMatchingCompare}, and
many others \cite{Ravesteijn10,Raykar08,Slabaugh10,Wang08}.

\begin{figure}[h]
\begin{center}
\epsfig{file=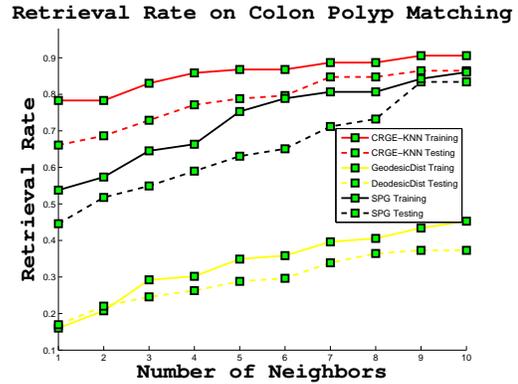, width=0.8\linewidth}
\caption{Retrieval comparison of using our proposed CTF method, the single-layer RVMMIL \cite{Raykar08}
classifier, and spectral projection on graph (SPG) \cite{Cai08} on colon polyp retrieval, in training and testing.}
\label{fig:RetrievalGeodesic}
\end{center}
\end{figure}

\subsection{Polyp retrieval}
 \label{sec:retrieval}

To fully leverage the topology-preserving property of learned $\mathcal{Y}$, we
also use it for polyp retrieval, which is defined as giving a query polyp in one prone/supine scan, to retrieve its counterparts in the other view.
To achieve this, we find the $k$ nearest neighbors ($k$NN) of a query $\mathbf{y}_i \in \mathcal{Y}$ using the classified polyps, and check whether the true match is inside the neighborhood of $k$NN. If the true matched polyp is in the $k$NN, a 'hit' will occur. We record the retrieval rate, as the ratio of the number of 'hit' polyp divided by the query polyp number, at different $k$ levels. Especially, high retrieval rate with small $k$ can greatly alleviate radiologists' manual efforts on finding the counterpart same polyp, with better accuracy. To show its advantage, we employ a traditional geometric feature based polyp retrieval scheme, namely geodesic distance that measures the geodesic length of a polyp to a fixed anatomical point (e.g., rectum), along the colon centerline curve. The retrieval rate comparison is illustrated in Fig. \ref{fig:RetrievalGeodesic}, for training and testing datasets. The results indicate that the retrieval accuracy can achieve $80\%$ when only $2$ to $4$ neighbors are necessary. This shows that
nonparametric $k$NN in $\mathcal{Y}$ subspace based retrieval significantly
improves the conventional polyp matching scheme, contingent on geometric
computation of geodesic distance and the SPG based retrieval. As a comparison, we also used the geodesic distance of polyp instances to do
retrieval. The geodesic distance measures the geodesic distance from the location of the polyp instance to a fixed point (the end of the colon). The retrieval rates of the tagged training and testing datasets are also shown in Fig. \ref{fig:RetrievalGeodesic},

\subsection{Lung Nodule Classification}\label{subsec:lung}

{\bf Data:} The lung nodule dataset is collected from $1000$ patients from
multiple hospitals in different countries, using multi-vendor scanners. Before
sample pruning, there are $28804$ samples of which $27334$ are negatives
and $1470$ are true nodule instances from 588 patients in training dataset. The
testing dataset contains $20288$ candidates, with $19227$ are negatives and $1061$ are positives of 412 patients. Several instances may correspond to the same lung nodule in one volume. All types of {\em solid, partial-solid and Ground Glass Nodules} with a diameter range of 4-30mm are considered. Each sample has 112 informative features, including texture appearance features (e.g. as the moments of responses to a multiscale filter bank, \cite{Ginneken01,Muhammad08}), shape (e.g. width, height, volume, number of voxels), location context (e.g. distance to the wall, at the right or left of the wall), gray value, and morphological features (e.g., obtained using the
edge-guided wavelet snake model as in \cite{Keserci02}).

\begin{figure}[t]
\centering
\subfigure[]{
\epsfig{file=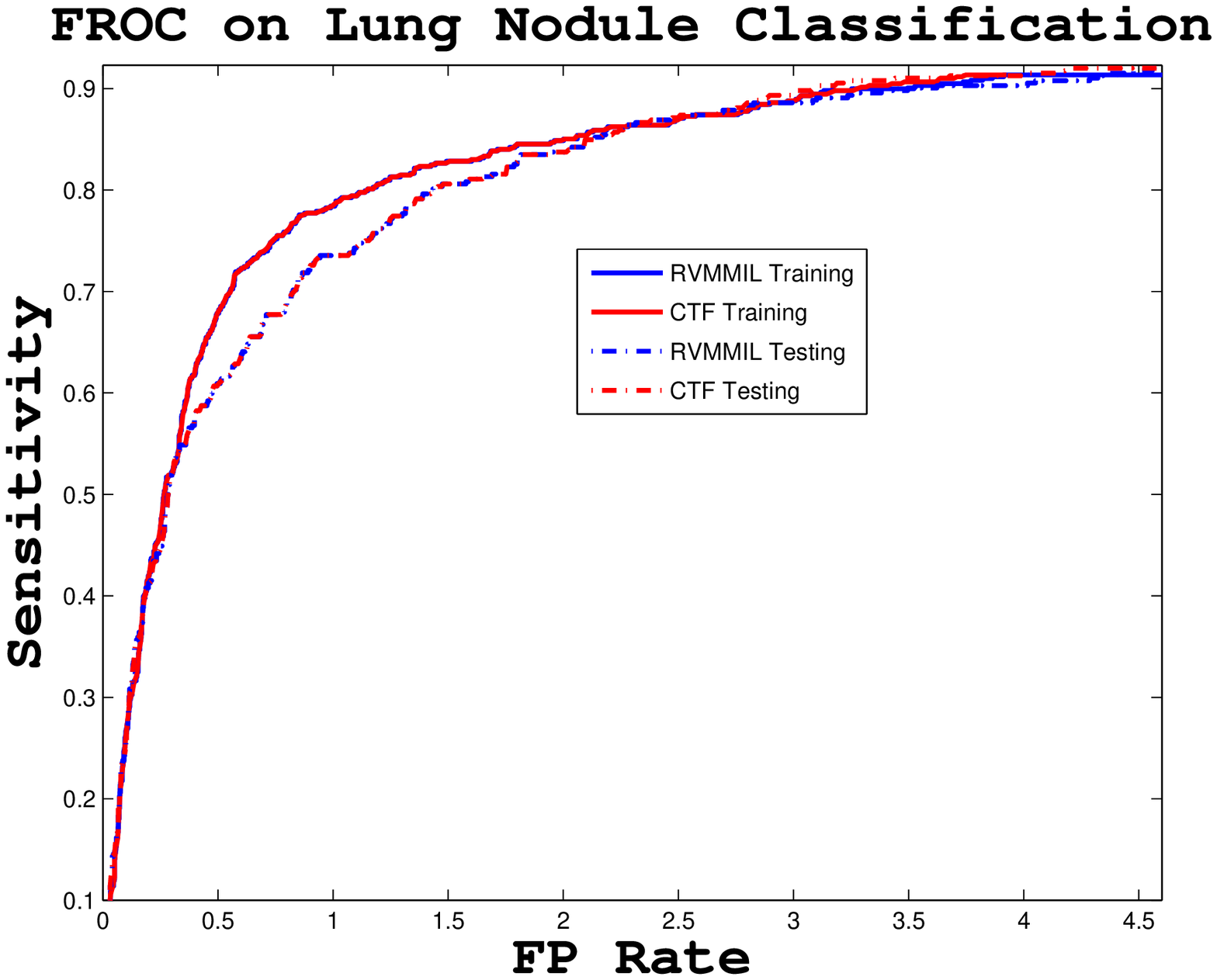, width=0.80\linewidth}
\label{fig:trainTestRVMMIL_CTF_lung}
}
\subfigure[]{
\epsfig{file=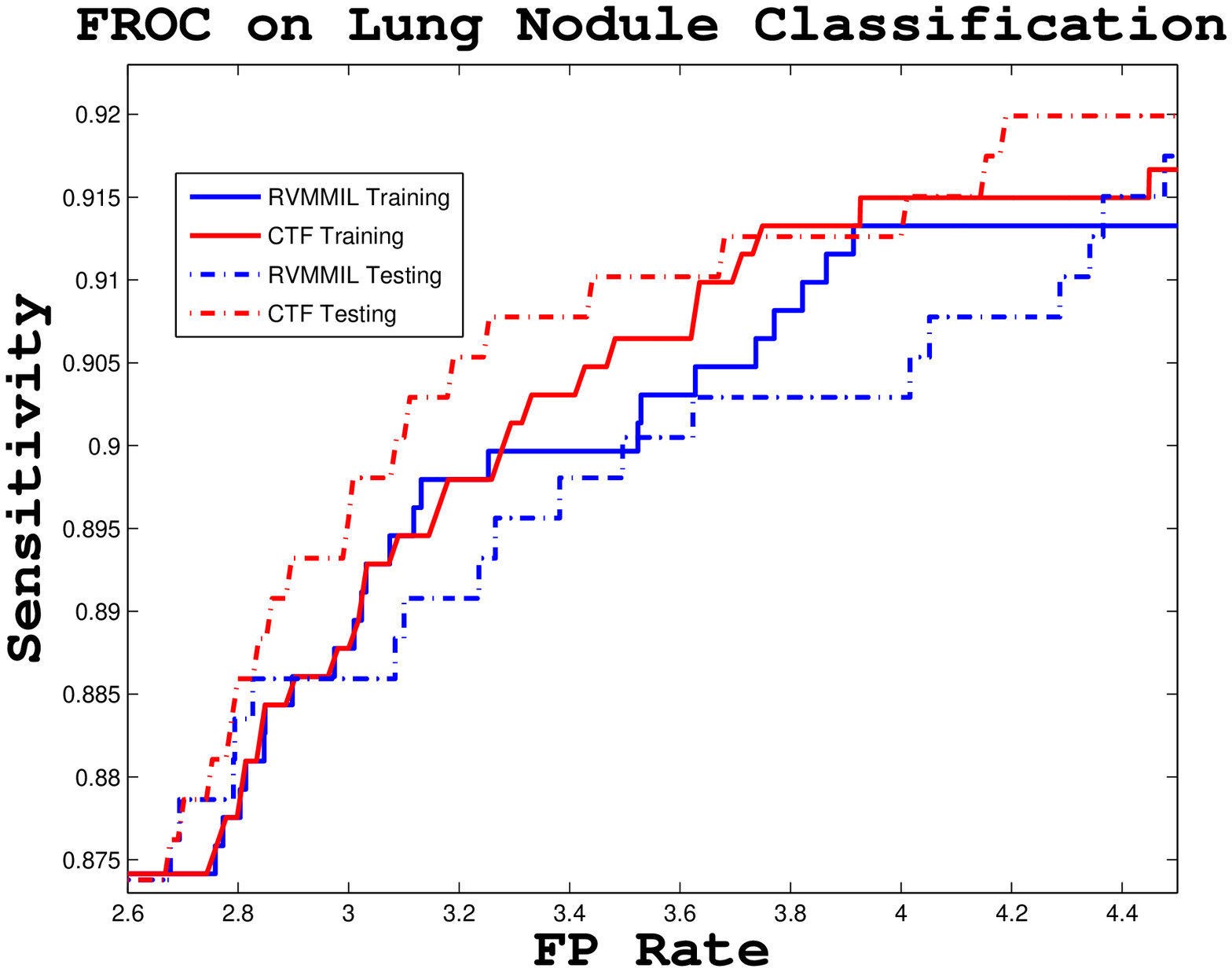, width=0.81\linewidth}
\label{fig:trainTestRVMMIL_CTF_lungZoomIn}
}
\caption{\subref{fig:trainTestRVMMIL_CTF_lung} FROC analysis using our proposed CTF method and RVMMIL classifier, in
training and testing of the lung nodule dataset. \subref{fig:trainTestRVMMIL_CTF_lungZoomIn} Zoom in of \subref{fig:trainTestRVMMIL_CTF_lung}
for the part of FP rate $\in[2.6,\,4.5]$.}
\label{fig:RVMMIL_CTF_lung}
\end{figure}

\begin{figure}[h]
\centering
\epsfig{file=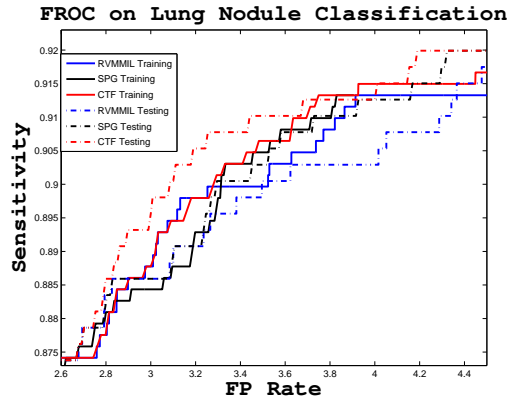, width=0.80\linewidth}
\caption{FROC analysis using our proposed CTF method, RVMMIL classifier and SPG in
training and testing of the lung nodule dataset.}
\label{fig:trainTestRVMMIL_CTF_SPG_lungZoomIn}
\end{figure}

First, FROC analysis of using our proposed coarse-to-fine classification
framework, compared with single-layer RVMMIL classifier, for the lung nodule
classification in training and testing is shown in Fig.
\ref{fig:RVMMIL_CTF_lung}. From the figure we can see that the testing FROC of CTF dominates the RVMMIL FROC, when the FP rate $\in[3,\,4]$, with $1.0\sim1.5\%$ consistent sensitivity improvements. We also compared with the SPG framework, and the FROC analysis is shown in Fig. \ref{fig:trainTestRVMMIL_CTF_SPG_lungZoomIn}. The comparison also shows
the higher classification accuracy
of our proposed method. Furthermore, our CTF classification performance compares favorably
with other recent developments in lung CAD \cite{Ginneken10}.

\begin{figure}[h]
\centering
\epsfig{file=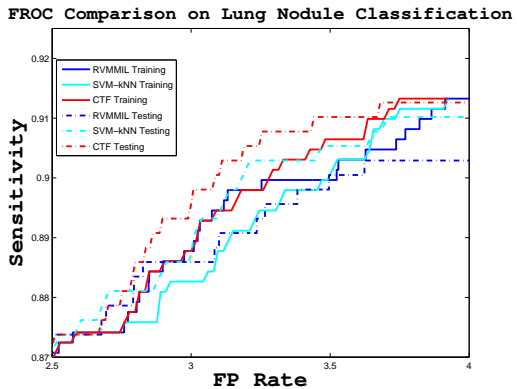, width=0.81\linewidth}
\caption{FFROC analysis using our proposed CTF method, RVMMIL classifier and SVM-$k$NN
classification scheme, in training and testing.}
\label{fig:SVMkNNTrainTestCompare}
\end{figure}

\begin{figure}[h]
\centering
\subfigure[]
{
\epsfig{file=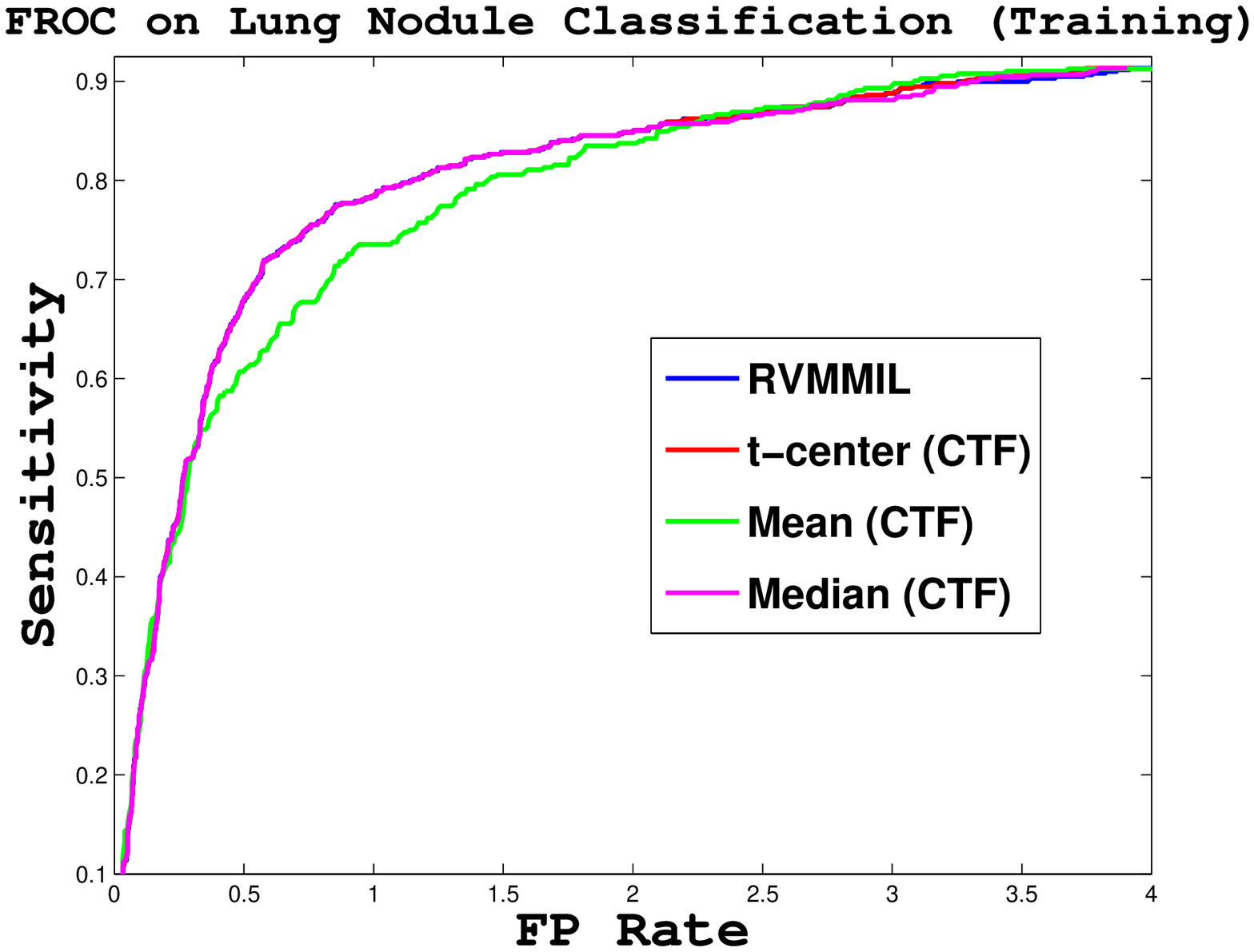, width=0.82\linewidth}
\label{fig:trainDifTemplateCompare_lung}
}
\subfigure[]
{
\epsfig{file=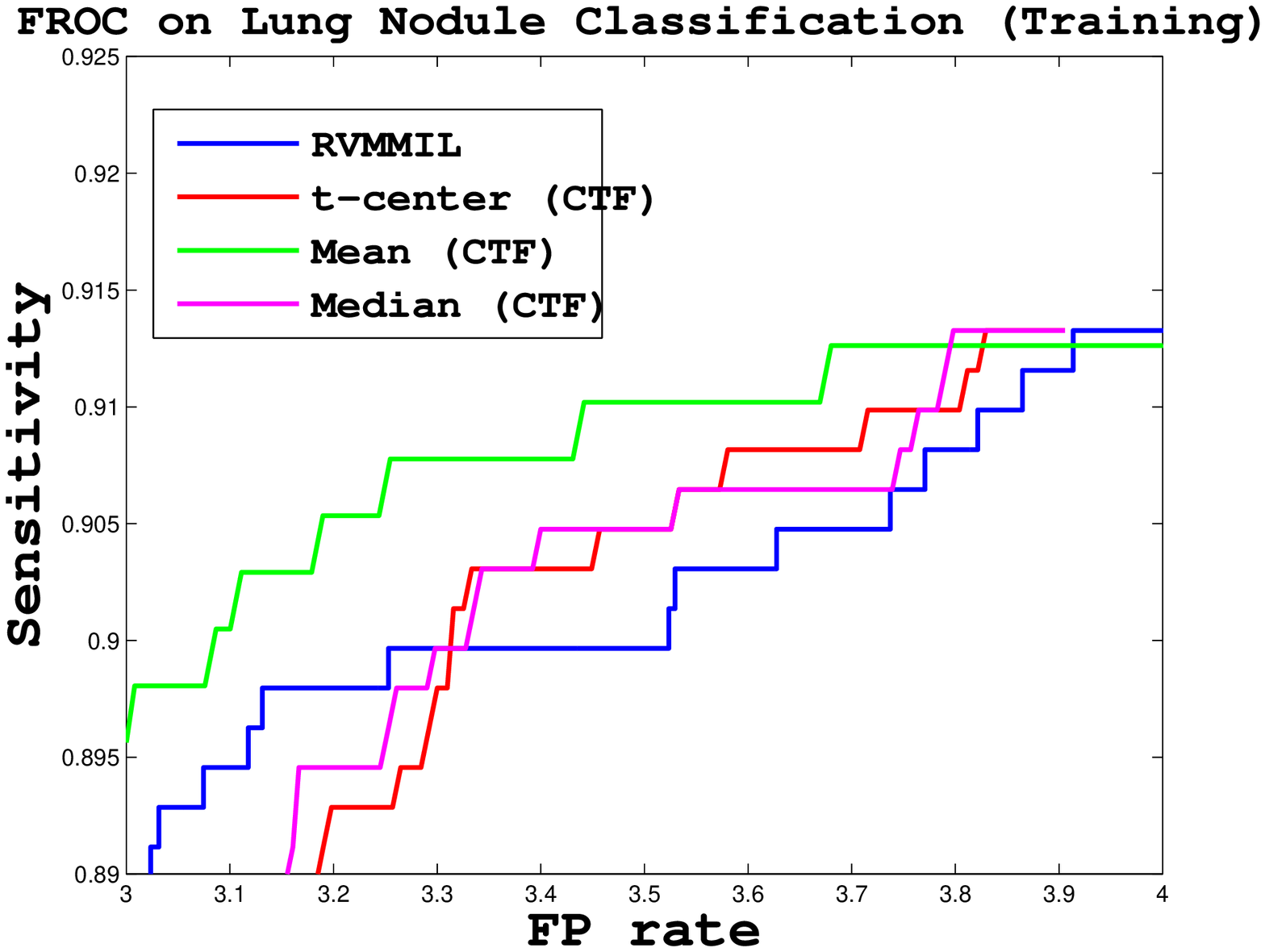, width=0.83\linewidth}
\label{fig:trainDifTemplateCompare_lungZoomIn}
}
\caption{FROC analysis using $t$-center, mean or median as estimated templates
in CTF, compared with RVMMIL classifier in training \subref{fig:trainDifTemplateCompare_lung}: original comparison
and \subref{fig:trainDifTemplateCompare_lungZoomIn}: after zooming in.}
\label{fig:difTemplatesCompareTrain}
\end{figure}

\begin{figure}[h]
\centering
\subfigure[]
{
\epsfig{file=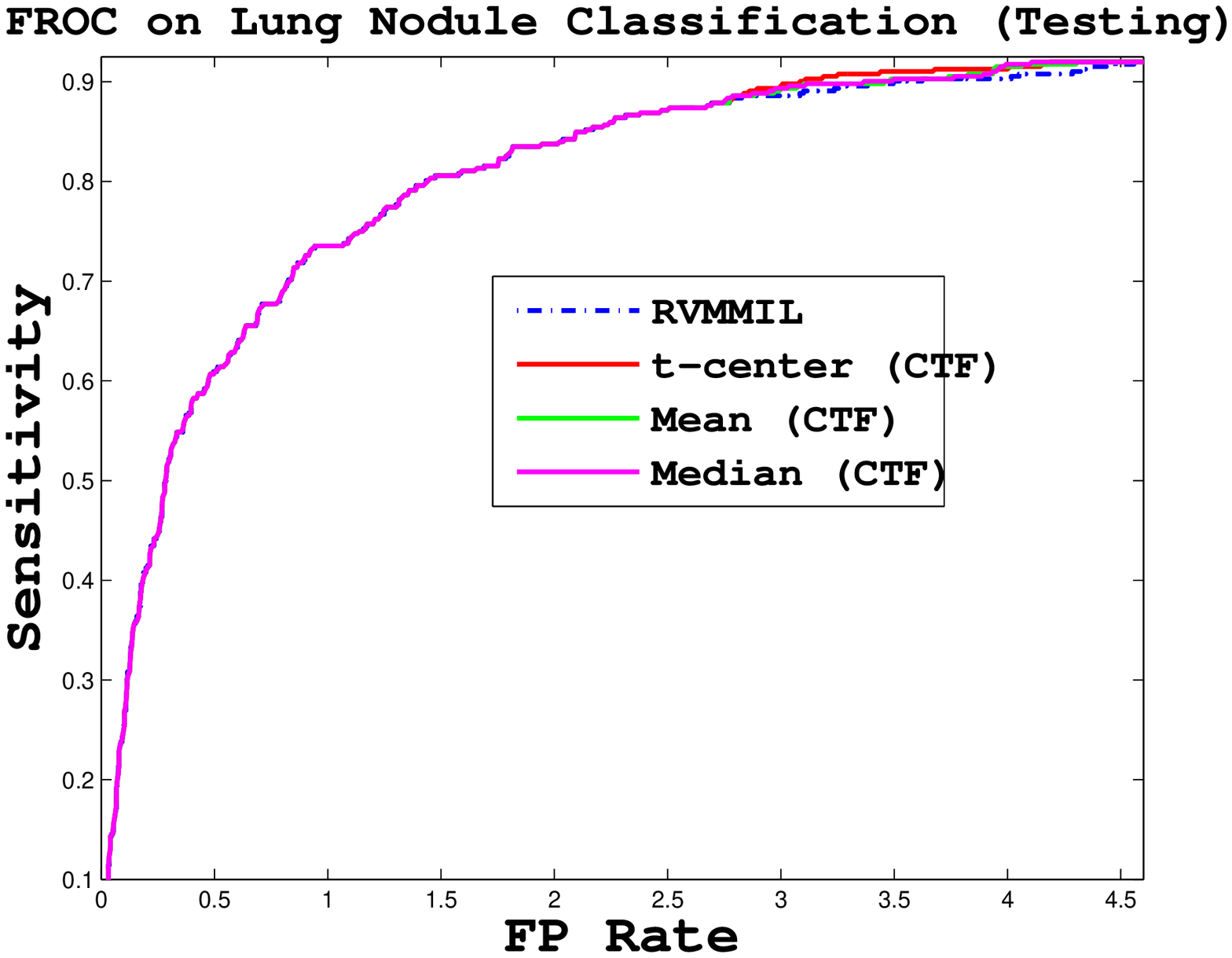, width=0.82\linewidth}
\label{fig:testDifTemplateCompare_lung}
}
\subfigure[]
{
\epsfig{file=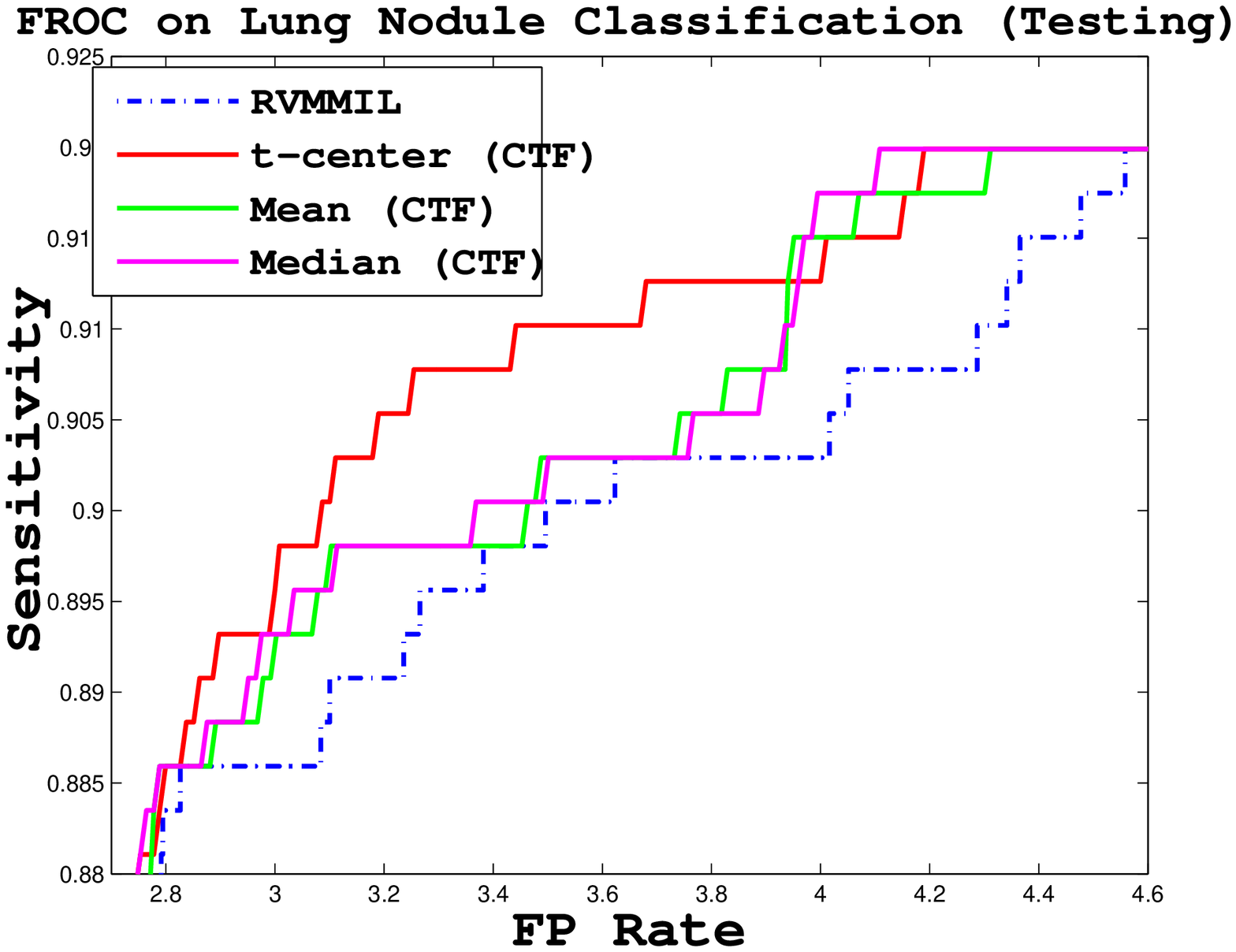, width=0.83\linewidth}
\label{fig:testDifTemplateCompare_lungZoomIn}
}
\caption{FROC analysis using $t$-center, mean or median as estimated templates
in CTF, compared with RVMMIL classifier in testing \subref{fig:testDifTemplateCompare_lung}: original comparison
and \subref{fig:testDifTemplateCompare_lungZoomIn}: after zooming in.}
\label{fig:difTemplatesCompareTest}
\end{figure}

Next we evaluate
the effects of using $t$-center (default), mean or median as estimated templates
in CTF. The comparison is shown in Fig. \ref{fig:difTemplatesCompareTrain} and Fig. \ref{fig:difTemplatesCompareTest}
on the training
and testing parts of the lung dataset. The comparison
 validates that $t$-center outperforms the templates formed by typical mean
or median method. Last, we compare our method with a related
locality-classification framework, SVM-$k$NN \cite{Zhang06} which shows highly
competitive results on image based multiclass object recognition problems.
SVM-$k$NN uses $k$NN to find data clusters as nearest neighbors and train a
support vector machine (SVM) on each locality group for ``divide-and-conquer''
classification \cite{Zhang06}. The comparison results are given in Fig. \ref{fig:SVMkNNTrainTestCompare},
showing that our method outperforms the
SVM-$k$NN method on both training and testing datasets.

\vspace{-1mm}

\vspace{-0mm}

%

\vspace{-0mm}

\vspace{-0mm}
%
%

\section{Conclusions \& Future Work}\label{sec:con}

Our main contributions are summarized in three folds. First, we
introduce a new coarse-to-fine classification framework for
computer-aided (cancer) detection problems by robustly pruning data
samples and mining their heterogeneous imaging features. Second, we
propose a new objective function to integrate the between-class
dissimilarity information into embedding method. Third, two
challenging large scale clinical datasets on colon polyp and lung
nodule classification are employed for performance evaluation, which
show that we outperform, in both tasks, the state-of-the-art CAD
systems \cite{Ginneken10,Murphy09,Ravesteijn10,Slabaugh10,Wang08}
where a variety of single parametric classifiers were used. For
future work, we plan to investigate optimizing the fine-level
classification in an associate Markov network \cite{Taskar04}
setting, which integrates structured prediction among data samples
(i.e., graph parameters are jointly learned with classification).

\bibliographystyle{abbrv}
\bibliography{noduleClassification}

\end{document}